\documentclass{sigkddExp}
\usepackage[utf8x]{inputenc}
\usepackage{amsmath}
\usepackage{cite}
\usepackage{lmodern,textcomp}

\usepackage{tikz}
\usetikzlibrary{matrix,arrows,decorations.pathmorphing}

\newcommand{\C}{\mathcal{C}}

\newcommand{\bb}[1]{\mathbb{#1}}
\newcommand{\st}{\ :\ }

\newcommand\restr[2]{{
  \left.\kern-\nulldelimiterspace 
  #1 
  \vphantom{\big|} 
  \right|_{#2} 
  }}

\begin{document}

\newcommand{\myunit}{1 cm}
\title{client2vec: Towards Systematic Baselines for Banking Applications}

\numberofauthors{2}

\author{
\alignauthor Leonardo Baldassini \\
       \affaddr{BBVA Data \& Analytics}\\
       \email{leonardo.baldassini@bbvadata.com}
\alignauthor Jose Antonio Rodríguez Serrano\\
       \affaddr{BBVA Data \& Analytics}\\
       \email{joseantonio.rodriguez.serrano@bbvadata.com}
}

\date{30 July 1999}

\maketitle

\begin{abstract}
The workflow of data scientists normally involves potentially inefficient processes such as data mining, feature engineering and model selection. Recent research has focused on automating this workflow, partly or in its entirety, to improve productivity. We choose the former approach and in this paper share our experience in designing the client2vec: an internal library to rapidly build baselines for banking applications. Client2vec uses marginalized stacked denoising autoencoders on current account transactions data to create vector embeddings which represent the behaviors of our clients. These representations can then be used in, and optimized against, a variety of tasks such as client segmentation, profiling and targeting. Here we detail how we selected the algorithmic machinery of client2vec and the data it works on and present experimental results on several business cases.
\end{abstract}

\section{Introduction}

Data science processes such as data exploration, manual feature engineering, 

validating hypotheses and constructing baselines can become convoluted, and often call for empirical and domain knowledge. Being a data science company, one of our concerns at BBVA D\&A is to provide our data scientists with tools to improve these workflows, in the spirit of works such as \cite{Duvenaud13, grosse2012, snoek2012, kanter2015, boselli2017ai, Lloyd2014}.

Most data analytics and commercial campaigns in retail banking revolve around the concept of behavioral similarity, for instance: studies and campaigns on client retention; product recommendations; web applications where our clients can compare their expenses with those of similar people in order to better manage their own finances; data-integrity tools.

The analytic work behind each of these products normally requires the construction of a  set of customer attributes and a model, both typically tailored to the problem of interest.

We aim to systematize this process in order to encourage model and code reuse, reduce project feasibility assessment times and promote homogeneous practices.

Our contribution to this end is client2vec: a library to speed up the construction of informative baselines for behavior-centric banking applications. In particular, client2vec focuses on behaviors which can be extracted from account transactions data by encoding that information into vector form (client embedding). These embeddings make it possible to quantify how similar two customers are and, when input into clustering or regression algorithms, outperform the sociodemographic customer attributes traditionally used for customer segmentation or marketing campaigns. We pursued a solution with minimal computational and preprocessing requirements that could run even on simple infrastructures. This is not damaging: baselines need to provide an informative starting point rather than an off-the-shelf solution, and client2vec helps generating them in few minutes, with few lines of code. Additionally, client2vec offers our data scientists the possibility to optimize the embeddings against the business problem at hand. For instance, the embedding may be tuned to optimize the average precision for the task of retrieving suitable targets for a campaign.

This paper describes our experience and what we learned while building client2vec; it is organized as follows: 

in Section \ref{sec:approach} we go through the principles we built client2vec on; in Section \ref{sec:previous_work} we mention relevant related work, from both the application and the algorithmic perspectives; in Section \ref{sec:transactions} we describe the structure of the account transactions data we used; in Section \ref{sec:methodology} we dig deeper into the algorithms we studied when developing our library, why we chose them and how we trained them; finally, in Section \ref{sec:experiments} we present experiments in the tasks of client segmentation, reconstruction of missing expenses and retrieval of targets for a commercial campaign.

\section{Our approach}

\label{sec:approach}

We built client2vec following an analogy with unsupervised word embeddings \cite{mikolov:2013, dhillon2012two,Chen:12, socher2011parsing}, whereby  
account transactions can be seen as words, clients as documents (bags or sequence of words) and the behavior of a client as the summary of a document. Just like word or document embeddings, client embeddings should exhibit the fundamental property that neighboring points in the space of embeddings correspond to clients with similar behaviors.

We thus identified two possible paths: one is to extract vector representations of transactions and compose them into client embeddings, as done with word embeddings to extract phrase or document embeddings via averaging or more sophisticated techniques \cite{socher2011parsing}; the other is to embed clients straight away. We explored the former option by applying the famed word2vec algorithm \cite{mikolov:2013} to 
our data and 
then pooling the embeddings of individual transactions into client representations with a variety of methods. For the latter approach, which is the one currently employed by client2vec, we built client embeddings via a marginalized stacked denoising autoencoder (mSDA) \cite{Chen:12}. For comparison and benchmarking 
purposes, we also tested the embedding comprising the raw transactional data of a client and the one produced by sociodemographic variables.

Embeddings are then turned into actionable baselines by casting business problems as nearest neighbor regressions. This builds on successful 
works in computer vision\cite{Torralba2008,hays2008,rodriguez2016} which adopt the  principle of the \textit{unreasonable effectiveness of data} \cite{halevy2009unreasonable}. As we will show, such an approach produces effective baselines in a variety of scenarios.

\section{Related work}
\label{sec:previous_work}

The need for automation has been noticed by the wider data science community. The AutoML Challenge proposed contestants to solve classification and regression tasks without human intervention \cite{guyon2015}. Projects like the Automatic Statistician\footnote{https://www.automaticstatistician.com/}, which aims at developing an ``artificial intelligence for data science'', for instance apply automatic, efficient exploration of the space of models for non-parametric regression by composing kernel structures \cite{Duvenaud13} or  matrix decomposition by organizing and exploring models in a context-free grammar \cite{grosse2012}. Other strategies involve using Gaussian Processes for hyperparameter search \cite{snoek2012} and autonomous feature engineering from raw data \cite{kanter2015}. Along the same lines,  automation may concern data cleaning \cite{boselli2017ai} or even summarizing data and models with natural language \cite{Lloyd2014}. A de facto standard for the rapid construction of baselines is the class of gradient boosting methods
\footnote{\protect{https://github.com/dmlc/xgboost/tree/master/demo\#machine-learning-challenge-winning-solutions}, accessed 2018-02-09.} \cite{friedman2001gbm}. 
However, we seek a method that can produce an unsupervised embedding of a client that be semantically interpretable, with neighboring points representing behaviorally similar clients, and that we can reliably employ in tasks as diverse as retrieval or clustering, as we do in Section \ref{sec:experiments}.

Consequently, we rather draw inspiration from the straightforward, successful application of textual embeddings produced by models like word2vec \cite{mikolov:2013} or Glove \cite{pennington2014glove} to data science problems such as predicting the outcome of soccer matches from Tumblr hashtags \cite{radosavljevic2014large} or improving the accessibility of the information in medical corpora \cite{minarro2014exploring}. More interestingly, mechanisms for embedding words are applicable to any indexable entity and such models have been extended to embed data other than text. This includes embedding nodes in a graph~\cite{figueiredo2017struc2vec}, items in recommender systems~\cite{barkan:2016item2vec}, Pinterest's pins \cite{Liu2017pin2vec} and even twitter users~\cite{benton2016user2vectwitter}, where different embeddings of the same entity are mixed using a form of canonical correlation analysis.

One drawback of these embedding methods is that they lack a formalism to embed sets or sequences of items, rather than the items themselves. On sequential data, methods such as RNNs offer in most cases state-of-the-art performance \cite{lipton2015critical}. However, these models are unsuitable to our use case: they are usually data-hungry and computationally burdensome, hence inadequate to rapidly and systematically build baselines. Autoencoders, on the other hand, have been shown to be effective at creating unsupervised embeddings of formal representations of sets (e.g. bag-of-items) ~\cite[Sec. 7]{bengio2013representation, Chen:12} and have thus found application, for instance, in embedding patient data~\cite{miotto2016deeppatient}. The latter work, in particular, leverages stacked denoising autoencoders (SDAs) and offers a very similar scenario to the ones we consider in our applications. We chose however to focus on mSDAs to reduce the computational cost associated with SDAs, Section \ref{sec:msdaembed}.

\section{\protect{Account transactions Data}}
\label{sec:transactions}

Current account transactions include movements like
direct debits, commission fees, coupons of financial products or debit card expenses. Said movements normally have an associated description, either in formatted 
or free text, and a movement code whose descriptivity is out of our control and ranges from the extremely general (a money transfer of any kind) to the very specific (the purchase of tickets of a given airline). The taxonomy of transactions is 
further enriched by assigning movement codes to one of 70 categories, e.g. \textit{Utilities} or \textit{Car insurance}. This operation is carried out by an internal tool employing search and NLP algorithms. Together with new structure, by miscategorizing some movements this procedure can also introduce noise, to which the algorithms we tested showed good resilience.

We chose to focus client2vec on an aggregation of current account data by client, year and transaction category. This aggregation can easily be applied to most commercial cases of interest, is readily available within BBVA and preserves enough information within a fairly succinct dataset. Past applications within BBVA include segmentation, campaigns, recommendations and automated finance management. Specifically, for a year worth
of data, we aggregate each client into a vector of 70 numbers, one per category, as depicted in Figure \ref{fig:raw_data}. Unless otherwise specified, this is the data format that we will refer to throughout the rest of the paper.

\begin{figure}[h!]
    \centering
    \begin{tikzpicture}[thick, scale=0.5, every node/.style={transform shape}]
    
    \tikzset{
        circlenode/.style={draw, color=red, text height=.2cm, anchor=east}
    }
    
    \matrix(dict)[matrix of nodes,
        nodes={align=right, anchor=east, text height=.2cm},
        row 1/.style={anchor=east, text height=.2cm, minimum height=.2cm},
        column 1/.style={nodes={align=right, anchor=east}}]{
        \hspace{1cm} & \hspace{.2cm} & \texttt{CAT1} & \texttt{CAT2} & 
        \texttt{CAT3} & $\cdots$ &  \texttt{CAT70}\\
        
        \texttt{client 1}   & \hspace{.2cm} & -10.15 & -527.11 & 1250.67 & $\cdots$ & - & \\
        \texttt{client 2}   & \hspace{.2cm} & - & -12.26 & 800.32 & $\cdots$ & 322.81 &\\
        \phantom{client 2} \\
        \texttt{client N}   & \hspace{.2cm} & -45.65 & -600.00 & 2400.56 & $\cdots$ & -241.67 &\\
    };
    \draw(dict-1-1.south west)--(dict-1-7.south east);
    \draw(dict-1-1.north east)--(dict-5-1.south east);
    \node[align=center] (dots) at (dict-4-1) {$\vdots$};

    \end{tikzpicture}
    \caption{Raw transaction data: a year worth of expenses aggregated
    averaged by client and expense category.}
    \label{fig:raw_data}
\end{figure}

\section{Tools and Methodology}
\label{sec:methodology}

As we remarked, at the core of the baselines client2vec allows to build, is an encoding of the raw transactions data. When building the client2vec library, we then had to compare various embedding strategies against relevant use cases. In this section we detail said strategies, motivate their choice and outline the
training methodology we used in our experiments.

\subsection{Sociodemographic variables}

The obvious fundamental benchmark to which we compared all methods
are sociodemographic variables: age, gender, income range, postcode, city and province. Such variables are typically considered by banks, retailers and other organizations for purposes like segmentations or campaigns. All of these variables are 
categorical, even the income, having been binned in several ranges. As such,
we one-hot encode them and then reduce the dimensionality of the
vector thus obtained in order to measure the Euclidean distance between two sociodemographic representations.

\subsection{Raw transactions}

The simplest behavioral embedding of a client could be expressed as ${\bf x_{\rm raw}} = (x_1, \ldots, x_k, \ldots, x_K )$, where $x_k$ is the aggregated expense of that customer in the $k$th category. This corresponds to one row of the table in Figure \ref{fig:raw_data}, possibly normalized (typically such that $\|{\bf x}_{\rm raw}\|_2^2 = 1$) or binarized.
This embedding helps us demonstrate that it is not only the use of transactional data instead of sociodemographic variables that 
can enhance the performance in certain banking applications, but also the
representation algorithms applied on the data.

Following the analogy with NLP: if an individual account operation in a certain category is interpreted as a ``word'', we can think of the raw 
transactions embedding as one-hot encodings (when binarized), or as ``bag of words'' (when continuous).

\subsection{Embedding via word2vec}

Word2vec \cite{mikolov:2013} is a family of  embeddings of words in documents, which express each word token with a dense vector. These vectors result from the intermediate encoding of a 2-layer network trained to reconstruct the linguistic context of each token and exhibit strong semantic properties, e.g. two nearby vectors refer to words that may share the same topic or even be synonyms. 

In order to apply word2vec to account transactions it was key to transform our transactions data into sequences of meaningful tokens that word2vec could process. Indeed, concepts like ``words'' and ``contexts'' do not 
arise naturally in financial transactions. Moreover, we need them
to reflect the different nature of 
each movement which the mere value of the transaction fails to capture: 1000€ spent on restaurants and 1000€ spent on
pension plans likely point towards two different behavioral profiles.

\subsubsection{Transactions quantization}
\label{sec:transquant}

To introduce a word-like entity we created a suitable quantization of our data. 
First, percentiles are computed for all transactions of our dataset in each category, i.e. computing percentiles for all the columns of the table in Figure \ref{fig:raw_data}. The percentiles are then used to define bins for the transactions of each category. Each bin is labeled with a word 
composed by the category identifier and the bin's boundaries

\footnote{Notice that prepending the category identifier is not strictly necessary
to disambiguate a bucket's category, since the probability of encountering two identical-boundaries bins from two different categories is 0.}. For example, referring to Figure \ref{fig:raw_data}, \texttt{CAT1\_-50.20:-7.16} is a possible  label of a bin containing the transactions in \texttt{CAT1} of both clients \texttt{1} and \texttt{N}. Such words then correspond to expenses in a certain category and range.  Finally, a quantized copy of the raw dataset is built by replacing individual transactions with the label of the bin they fall inside, hence yielding a finite set of repeating words that depend on the nature of the transaction. The dataset thus transformed assumes some of the structural features of natural text, such as a finite vocabulary and a sentence-like organization.

A limitation of this approach is that properties like syntactical structures or parts of speech are here completely absent. Although a definition of context as co-occurrence of bin labels still holds, it is however be much weaker than that existing in text since our data, in the aggregation we are using, lacks a form of ordering. We found out that this severely hinders the performance of word2vec embeddings, which encouraged us to explore the alternative offered by mSDAs, Section \ref{sec:msdaembed}.

\subsubsection{From transactions to clients}

Applying word2vec on the quantized dataset will yield transaction embeddings, which we 
then need to compose into client embeddings. Strategies 
such as concatenation or mean-pooling have been proposed in the context of computational linguistics \cite{bengio2003neural, le2014distributed}. While intuitive
and computationally inexpensive, they present weaknesses such as
overly increasing the dimensionality of the representation or losing 
all information about word order. The computer vision community also 
extensively investigated the problem of pooling local representations \cite{boureau2010theoretical, murray2014generalized}
leveraging ideas such as Fisher kernels
\cite{jaakkola1999exploiting, perronnin2007fisher}, max-pooling \cite{boureau2010theoretical} or geometric approaches such as
VLAD \cite{jegou:vlad}. The latter, in particular, if applied to textual data, would amount to computing cluster centroids for the word vectors
and subsequently 
concatenating the differences between each word vector and the corresponding centroid for all the words in a sentence. Considering that the centroids can be seen as ``representative'' words, VLAD allows to embed a sentence by its proximity to representative concepts.

This makes sense even in the context of transaction embeddings, where we'd
like a client embedding to capture the salient behavioural characteristics
of an individual. We therefore composed our low-level transaction embeddings obtained through word2vec with mean-pooling as well 
as VLAD.

\subsection{Embeddings via mSDAs}
\label{sec:msdaembed}

As anticipated in Sections \ref{sec:previous_work} and \ref{sec:transquant}, after observing the inconsistent performance of word2vec-based embeddings, following the line of work of \cite{miotto2016deeppatient} we turned our attention to autoencoders. In 
particular, we chose to employ a marginalized stacked denoising autoencoder \cite{Chen:12} 
which offers several advantages: ease of implementation and deployment; modest 
computational cost; excellent performance in domain adaptation tasks; and
a natural manifold-learning interpretation explaining why neighboring 
embeddings represent clients with similar behaviors \cite{Chen:12, vincent2008extracting}. We see all of these as valuable properties in a baseline.

MSDAs' starting architecture is that of denoising autoencoders (DAs)  \cite{vincent2008extracting}, which are trained to minimize 
a reconstruction error such as the squared error
$L(\mathbf{x}, \mathbf z) = \|\mathbf x-\mathbf z\|^2
$
used for continuous-valued data. In DAs the input $\mathbf x$ is first corrupted with noise, 
yielding a $\bar{\mathbf x}$ which is then encoded as 
$\mathbf y \doteq f(x) = \sigma(W\bar{\mathbf x} + \mathbf b) $ 
and finally decoded back to 
$\mathbf z \doteq g(y) = \sigma(W^\prime \mathbf y + \mathbf b^\prime)$, as usual 
with weight matrices $\mathbf{W}$, $\mathbf W^\prime$, biases $\mathbf b$, $\mathbf b^\prime$ and squashing nonlinearity $\sigma$. Denoising autoencoders can be stacked, yielding increasingly richer
representations of the input data \cite{vincent2008extracting, vincent2010stacked}.

The aim 
of this architecture is to let the autoencoder learn the statistical dependencies among the variables in 
$\mathbf x$, as that will allow it to reconstruct the value of a variable given the others. This is the information which an autoencoder encodes in the embeddings it produces. Therefore, two clients whose embeddings are close together don't necessarily have the same expenses, but rather expenses with similar statistical dependencies.
Following the manifold-learning and the generative-model interpretations of DAs \cite{vincent2008extracting} we can then think of the autoencoder as a way to learn a manifold on which neighboring points are mapped back to behaviorally similar client by the decoding function $g$. Furthermore, the good performance shown by autoencoders
in transfer learning and domain adaptation 
 \cite{Chen:12, zhuang2014transfer_autoenc, zhuang2015supervised_transfer_autoenc} suggests a statistical robustness suitable to build versatile baselines.

The change introduced by mSDAs is to marginalize out the noise distribution and train the autoencoder as if it had 
observed infinitely many noisy variations of the training set. The
marginalized autoencoder is then modeled as a single reconstruction matrix $\mathbf M\in\bb R^{d \times d}$, with $d$ the dimension of the input. Following \cite{Chen:12}, when the number of noisy observations goes to infinity, $\mathbf M$ can be written
as
\begin{equation}
\label{eq:mda_W}
\mathbf M = \bb E [\mathbf P] \bb E [\mathbf Q]^{-1}\ ,    
\end{equation}

where $\mathbf P$ is the cross-correlation matrix between clean and noisy data, 
$\mathbf Q$ is the precision matrix for the noisy observations, and the expectation is taken over the noise distribution. By rewriting  \eqref{eq:mda_W} as $\mathbf M \bb E[\mathbf Q] = \bb E[\mathbf P]$ it is possible to interpret the weights in $\mathbf M$ as follows: higher weights indicate which features in the data carry most information
when reconstructing a target feature. 

This construction of the autoencoder saves us from specifying the dimension of the 
intermediate encoding, while we use the the ``reconstructed'' version 
$\sigma(\mathbf M \mathbf x+\mathbf b)$ as the representation of $\mathbf x$ which, in the non-stacked case,
has the same dimension of $\mathbf x$. Such embedding therefore has the following interpretation: each entry holds the most likely reconstruction value for that variable. In 
the context of our data, this means that each entry holds the most likely average 
transaction value in the corresponding category, given all the other transactions of that 
client.

A limitation and at once an advantage of mSDAs is that their implementation boils down to solving the system of linear equations specified in \eqref{eq:mda_W}, yielding a complexity of $O(d^3)$, that is cubic in the number of features
\footnote{Notice however that the model can be adapted to deal with high-dimensional data \cite{Chen:12}.}. Nevertheless, we did not experience any performance issue on the datasets we consider in Section \ref{sec:experiments}. These sets reflect those normally used for some commercial applications within BBVA, containing between 120K and 300K clients, each represented with 70 to 700 features. \mbox{MSDAs} are therefore an appropriate choice for our use cases, easy to implement, hassle-free to deploy and fast to train 
and tune due to their small numbers of hyperparameters. Indeed, the only model-related
hyperparameters in our training are the probability $p$ for the Bernoulli masking noise and the 
number of stacked layers. One may also treat the choice of the nonlinearity $\sigma$ as a hyperparameter, but we chose to keep a $\tanh$ for all our models.

Finally, in terms of preprocessing, mSDAs' requirements are also minimal: we can use the raw transactions table with just a few adjustments such as 
binarizing, $L_2$-normalizing, log-normalizing, max-normalizing or rescaling from $[0,1]$ to $[-1, 1]$. These are also treated as 
hyperparameters and optimized for each individual use case the mSDA is being 
employed on.

\subsection{Model selection}
We treat the preprocessing options for mSDAs listed above like hyperparameters to optimize at train time. Likewise, the hyperparameters for the word2vec benchmark are the word-embedding dimension and the context window size \cite{mikolov:2013}, while for the raw transaction embeddings we only choose whether to $L_2$-normalize, log-normalize or binarize. The optimization is carried out separately for each use case we consider.

\section{Experiments and Results}
\label{sec:experiments}

In this section we report the experiments we ran to check which 
of the client representation methods was
the most versatile.

In particular, we first evaluated the embeddings against two base use cases: client
segmentation and the reconstruction of the presence of transactions in a left-out category. The former also provided us with a qualitative description of 
how mSDA-based embeddings can extract meaninfgul behavioral patterns 
from transactional data. From these experiments it became apparent that mSDAs were offering the best representations, so we finally tested them on a commercial case
of campaign targeting. 

Observe that the use cases we are considering may be better tackled with more sophisticated machinery, e.g. advanced clustering techniques \cite{filippone2008surveyclust} in Sections \ref{sec:clustering} and \ref{sec:typical} or recommender systems \cite{bobadilla2013recommender} in Sections \ref{sec:missing_cat} and \ref{sec:targeting}. However, our aim is not to provide a state-of-the-art solution to these use cases, but rather to employ them as tests to select a good embedding for client2vec and, finally, to show the potential of the baselines thus obtained. All the experiments have been carried out on anonymized data and with sensitive information removed or adequately tokenized.

\subsection{Clustering of clients}
\label{sec:clustering}

One of the most immediate use cases for a client embedding is generating a segmentation of the 
client base. One example of such segmentation arises in the functionality ``Compare yourself'' of our banking app
\footnote{www.bbva.es/eng/general/bconomy/pedro/index.jsp, accessed 2018-02-09.}. Here, customers can compare their own expenses in certain categories against those of their sociodemographic segment, which is defined by 4 variables (age, income, gender and postcode). Note that these attributes have been selected and constructed by manual adjustment over time.  For instance, an attribute like ``income" is prone to be built progressively by trial and error as there are many non-deterministic signals of income in our data, which a data scientist might only discover progressively. 

Our proposal for this first use case is to plug the embeddings we're testing into a clustering algorithm and compare the homogeneity of the values of a left-out target category in the resulting clusters. This approach fits the interpretation of mSDAs highlighted in Section \ref{sec:msdaembed}.

\subsubsection{Setup and evaluation}

We use a dataset with a year worth of data for 120k clients. For each target category $t$ and embedding $E$, calling $x_t$ the expense of client $\mathbf x$ in category $t$ we use  
the vector 
$$\mathbf x_{-t} = (x_1, \ldots, x_{t-1}, x_{t+1}, \ldots, x_K)$$ 
of remaining attributes to compute 
an embedding $E(\mathbf x_{-t})$. That is, each time we leave out one target category for evaluation purposes.  The embeddings thus obtained are clustered via K-means in 300 clusters, a number comparable to that of the sociodemographic segments. While better clustering algorithms are available \cite{ filippone2008surveyclust}, K-means is a common first choice in the phase of baseline construction.

Call $\mathcal T$ the set of transaction categories, $\C_t(E)$ the set of clusters produced for category $t\in\mathcal T$ by the embedding $E$, $\mathcal{S}_{c} = \{\mathbf x \st E(\mathbf x_{-t})\in c\}$ the set of customers whose embedding is in $c\in \C_t(E)$, and $\mathcal{S}^{{\rm out}}_{c} = \{x_t \st \mathbf x \in \mathcal S_c\}$ the corresponding values in the left-out category. We define the dispersion of $\C_t(E)$ by the median spread within each cluster $c$, with the spread being the standard deviation ${\rm std}(\mathcal{S}^{\rm out}_c)$. By repeating this for all $t$, we can evaluate embeddings by the average dispersion across all categories: 
\begin{equation}
\Delta(E) = \frac{1}{|\mathcal{T}|}\sum_{t\in \mathcal{T}} {\rm median}\big[\{{\rm std}(\mathcal{S}^{\rm out}_c)\}_{c\in \C_t(E)}\big]
\label{eq:cluster:eval}
\end{equation}

The rationale is that if customers are really similar within a cluster, this similarity should generalize to the unseen category and produce a low dispersion.

\subsubsection{Results}

 Table \ref{tab:cluster:results} shows the results of the experiment: the nominal values of $\Delta(E)$ and variation relative to the sociodemographic embedding baseline. Simply moving away from the use of sociodemographic variables
already produces a significant improvement, but within the embeddings built on transactional data the one offering the best
performance is that produced by an mSDA.

To us, this is a critical implication in terms of data science productivity: for this use case, a baseline requiring one call to the client2vec library to obtain client embeddings, plus K-Means, yields more homogeneous customer segments than those of hand-crafted, domain-specific attributes. To complement these quantitative results, the next section will provide a qualitative assessment of whether mSDAs really identify customers with similar behavior.

\begin{table}[h!]
\centering
\begin{tabular}{lccc}
\hline
Method & Relative $\Delta$ & Nominal $\Delta$ \\
\hline

Sociodemographic & +0.0 & 266e-5 &\\
Transactions & +58.66 & 110e-5 & \\
mSDA & {\bf +61.62} & {\bf 102e-5} &\\
word2vec + Avg & +48.19 & 137e-5 &\\

word2vec + VLAD & +50.59 & 131e-5 &\\
\hline
\end{tabular}
\vspace{5pt}
\caption{Relative and nominal values of our dispersion metric $\Delta$ for
various embedding methods.}
\label{tab:cluster:results}
\end{table}

\subsection{Case study: the typical traveler}
\label{sec:typical}

We 
consider the case of clients that have travel-related 
expenses, and conjecture that a typical traveler should also 
have expenses in the {\em Hotels} category. This 
mining experiment further illustrates the benefits of transactional 
embeddings over sociodemographic variables.

We proceed as in Section \ref{sec:clustering}, using {\em Hotels} as target category. In addition to mSDAs and sociodemographic variables, here we also consider embeddings obtained by assigning a person to a random sociodemographic segment, e.g. by randomly changing their income or age ranges. We aim to extract different definitions of {\em typical travelers} according to different embeddings. To do so, in each embedding space we look at the 10 most central clients in each of the 10 densest clusters. The clients considered in different embedding spaces may differ. The proximity of a client 
to their cluster's centroid represents the typicality of that client 
within the cluster, while cluster density is a measure of feature uniformity of the a cluster.
Looking at different dense clusters, finally, prevents us from focusing exclusively on a single behavioral pattern or sociodemographic segment.

\begin{figure} [h!]
\begin{center}
\includegraphics[width=.36\textwidth]{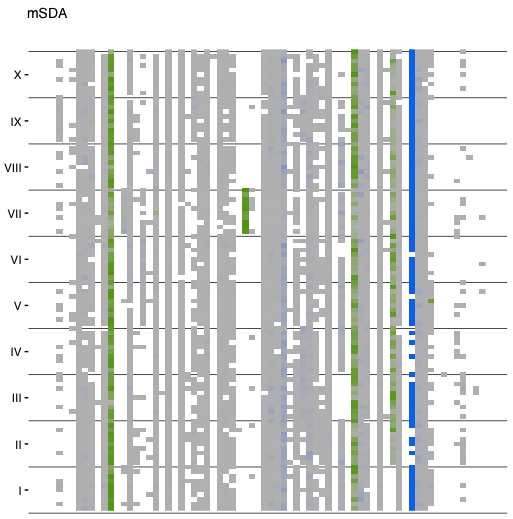}
\includegraphics[width=.36\textwidth]{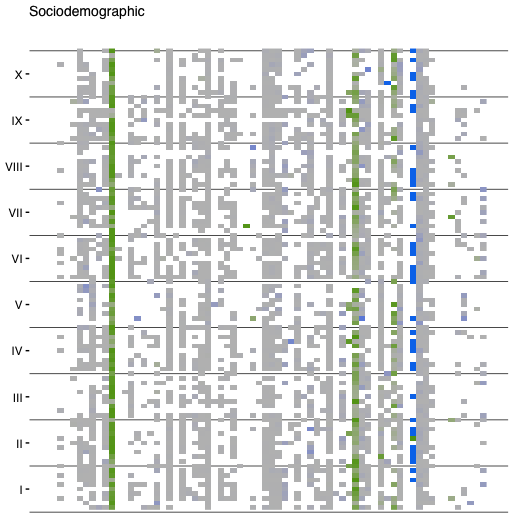}
\includegraphics[width=.36\textwidth]{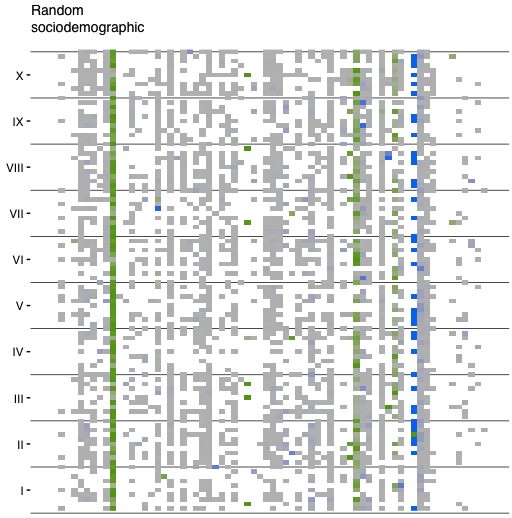}
\end{center}
\caption{(Best viewed in color) Transaction patterns of typical clients with travel expenses. Each matrix corresponds to a different embedding. Blocks of rows correspond to clusters, rows to clients and columns to transaction categories. Grey dots mark expenses, green ones incomes and the blue column the expenses in hotels.}
\label{fig:travel_case}
\end{figure}

Figure \ref{fig:travel_case} compares the transaction patterns of 
these typical clients. 
Each matrix corresponds to a different embedding. The Roman-numbered blocks of rows correspond to clusters, rows to clients and columns to transaction categories. Green dots correspond 
to incoming transactions (e.g. salary payments), while grey dots to outgoing ones. The blue column corresponds to expenses in the {\em Hotels} category.

The figure highlights how typical clients in the space of mSDA embeddings actually have very similar expense patterns (solid blocks of grey). 

Furthermore, most of them have expenses in the {\em Hotels} category, confirming our initial hypothesis. Interestingly, even the clients not having an expense in {\em Hotels} roughly follow the grouping in clusters, being present mostly in clusters II and IV.

All this structural information is lost with the 
sociodemographic embedding, where the spending behavior 
appears fragmented and uncorrelated to the cluster 
structure. In fact, these patterns are virtually indistinguishable from those extracted with random sociodemographic segments.
This suggests that sociodemographic variables capture close to no information about people's spending habits.

\subsection{Missing category prediction}
\label{sec:missing_cat}
Another measure of the effectiveness of a client embedding is being able to reconstruct 
whether a person had an expense in a target category. As some categories can be interpreted as families of products or services, e.g. \textit{Saving products}, \textit{Bars and restaurants} or \textit{Sports clubs}, this use case can be seen as an approximation of product propensity prediction.  In fact, it 
paves the way to the more commercially relevant experiment described in 
Section \ref{sec:targeting}, in which we'll look for clients that 
\textit{should} have an expense in a category, but don't.

\subsubsection{Setup and evaluation}

Here too we use a small dataset of 120K clients, splitting it in 100K
customers as the training set, 10K as a validation set and 10K as the test
set. We follow the notation of Section \ref{sec:clustering} and define 

\begin{equation}
    \label{eq:theta}
    \theta_t(\mathbf x) = |\textrm{sgn}(x_t)|\ ,
\end{equation}

which indicates whether client $\mathbf x$ has a transaction in the target category $t$. To estimate it, we 
perform nearest neighbor regression in the embedding space. Call 

\begin{equation}
    \label{eq:neighborhood}
    \mathcal N^E_k(\mathbf x) = \{k \textrm{ nearest neighbors to }
    \mathbf x\}
\end{equation}

where the neighbors are defined by a dot-product similarity in the space of the embedding $E$. With this set we estimate

\begin{equation}
    \label{eq:theta_hat}
    \hat \theta_t(\mathbf x) = 
        \dfrac{1}{k} \sum_{\mathbf z\in \mathcal N^E_k(\mathbf x)} 
            \theta_t(\mathbf z) 
\end{equation}

obtaining a preference score for the event that client $\mathbf x$ had 
an expense in category $t$.
Drawing from the information retrieval methodology, we evaluate our predictions by average precision (AP) and
precision at 100 (P@100), which for clarity we recall here. For a target 
category $t$, we compute:

\begin{equation}
\label{eq:AP_t}
    \textrm{AP}_{\theta_t}(\mathcal X) = \frac{\sum_{\mathbf x \in \mathcal X} P(\hat \theta_t(\mathbf x))\cdot\theta_t(\mathbf x)}
    {\sum_{\mathbf x\in \mathcal X} \theta_t(x)}\ ,
\end{equation}

where the sum is taken over the set $\mathcal X$ of all clients ordered by decreasing values of $\hat \theta$ and $P(\hat\theta_t(\mathbf x))$ is the precision computed on those clients with score higher than $\hat \theta_t(\mathbf x)$.

The P@100 is just the precision computed for the top-100
ranked predictions. The results discussed below are taken averaging the $\textrm{AP}$ for 15 representative
categories that bear commercial relevance
for the bank or the client, e.g. \textit{Expense in groceries} or \textit{Contributions to
saving products}.

\subsubsection{Results}

We evaluate results as in Section \ref{sec:clustering}, but in the case of raw transactions and mSDA we look at 
both continuous and binarized transactions separately. Notice that in 
applications this preprocessing option would be optimized as a 
hyperparameter. We choose not to do this here to show that mSDAs are all but 
unaffected by such a relevant change, while the raw transactions 
embedding greatly benefits from it.

The results are presented in Table \ref{tbl:binary}.
Here mSDA embeddings still maintain an 
advantage over the alternatives, although to a lesser extent than in the segmentation use case. Here too, however, the improvement over sociodemographic variables is significant, being above 54\
Notice also the poor performance of word2vec-based methods, which had offered 
much better results in Table \ref{tab:cluster:results}. This is probably to be attributed to the lack of temporal information in our dataset and the coarseness of transaction categories, which make the concept of a local context of a word somewhat weak, if not ill-posed, as we remarked in Section \ref{sec:transactions}. As a result, on tasks more complex than clustering, the performance of word2vec embeddings may be hindered.

Interestingly, the performance of the raw transactions embedding 
greatly benefits from the change from binary to continuous data while
mSDAs seem to be left largely unaffected - in fact, their performance deteriorates a little. We conjecture that this may indicate that the autoencoder 
doesn't need the information provided by the average transaction values when
simply predicting {\em whether} an expense occurred, being 
the correlations among the variables in binary input data 
just as informative to that end. On the other hand, building exclusively on the transaction values,
the raw transactions embedding is affected by the extreme coarseness of the binarized data.

\begin{table}[!h]
\centering
\begin{tabular}{lcc}
\hline
Method &  AP & P@100 \\
\hline
\hline
Sociodemographic  		& 0.378 & 0.533\\
\hline
Transactions (bin) 	& 0.324 & 0.466\\
mSDA (bin) 				& {\bf 0.584} & {\bf 0.771}\\
\hline
Transactions (cont) 	& 0.558 & 0.704\\
mSDA (cont) 				& 0.579 & 0.770\\
word2vec + Avg 		& 0.269 & 0.359\\
word2vec + VLAD 	& 0.274	& 0.380\\
\hline
\end{tabular}
\caption{\label{tbl:binary}Results of NN regression for missing category prediction. The AP improvement of mSDA 
embeddings over sociodemographic variables is 
above 54\%}
\end{table}

\subsection{Campaign targeting}
\label{sec:targeting}

In the experiments above we've seen how mSDAs achieved a consistently better performance in clustering and regression contexts, suggesting this is the first off-the-shelf embedding our colleagues should try from our library when starting to investigate a new mining task or commercial application. To reinforce this point, in this section we show how mSDA-based embeddings can be employed to create, 
with minimal effort, a commercially valuable method to retrieve sensitive targets for 
a commercial campaign. Application of non-trivial data-analytic methods for segmentation 
and targeting is not new, but effective methods like the one proposed in \cite{hsieh2009enhancing} usually entail a significant technical debt in development and maintenance, which becomes negligible with our library.

This experiment provides additional evidence that 
embeddings on transactional data have the potential to speed up data mining and analysis and even produce 
readily consumable solutions which can then be iterated upon, if needed.

\subsubsection{Car and motorcycle insurance}

One of the transaction categories we use in our data is that of a payment towards a car or motorcycle insurance, which creates a dataset of behaviors of the people who
purchased that product (our positive samples). For this experiment we are using a richer dataset than the ones considered before, 
namely losing the category-wise aggregation in favour of movement-descriptor level data.
Notice that this is a further simplification of the preprocessing we apply.
We cast our search for similar clients 
as an IR problem \cite{frakes1992information, liu2009learning2rank}, and in particular we consider a randomly sampled set $\mathcal Q$ of $Q=1000$ {\em seed} positive clients, also referred to as {\em query} clients, who have a car insurance with BBVA, and a \textit{database} $\mathcal D$ of $D=3E5$ clients where only 7000 individuals also have the same car insurance. 

We proceed precisely as in Section \ref{sec:missing_cat} by embedding via mSDA and then predicting using a set of $k$ nearest neighbors computed with dot-product similarity. The
choice of $k$ is essentially business-dependent: the more neighbors we consider 
for each query, the more sensitive targets for our campaign we're likely to retrieve. 
However, as it is to be expected, beyond a certain threshold for $k$ the precision of this method will start deteriorating. In fact, it can be reasonably 
argued that using a small $k$ will produce 
only a very restricted target set, given that the query customers supposedly share similar properties, hence they may also share their nearest 
neighbors - especially with a randomly sampled $\mathcal Q$. In practice, that is certainly the case, but at the end of this section we will show that, in this instance, it is possible to achieve a satisfactory trade-off between diversity and precision.

Following Section \ref{sec:missing_cat}, we evaluate the efficacy of our approach using the mean average precision at $k$ (MAP@$k$), which we define below
\footnote{$k$ is the same variable defining the number of neighbors.}.
We use the definition in \eqref{eq:neighborhood}, assuming here that $E$ is the mSDA embedding and that the neighborhood is computed within $\mathcal D$ alone. We then define

\begin{equation}
    \textrm{MAP@}k = \frac{1}{Q}\sum_{\mathbf q\in\mathcal Q}
           \textrm{AP}_{\mu_M}(\mathcal N^E_k(\mathbf q))\ ,
\end{equation}

where $\textrm{AP}_{\mu_M}(\mathcal N^E_k(\mathbf q))$ is the average precision defined in \eqref{eq:AP_t} computed on the set of neighbors of $\mathbf q$. The difference here is that we replace the indicator $\theta$ of \eqref{eq:theta} by the value $\mu$, defined as follows.  Given $M$ a subset of movement descriptors
and $\mathbf x$ a vector of a client's transactions aggregated by movement descriptor, 
$$
\mu_M(\mathbf x) = \textrm{sgn}\left(\sum_{m\in M} |x_m|\right)
$$
defines the relevance (positivity)  of $\mathbf x$. That is, $\mathbf x$ is a true positive, hence relevant to the retrieval task, if it has a transaction for any of the movement descriptors in $M$. The extension of the estimator in \eqref{eq:theta_hat}  as $\hat \mu_M$ naturally follows. The APs are then
averaged over the queries to compute the MAP@$k$.

\begin{table}[!h]
\centering
\begin{tabular}{llccccc}
\hline
Method              & MAP@50    & @2E4      & @5E4      & @9E4  \\
\hline
mSDA + dot 	        & 0.000    	& 0.061	    & 0.055	    & 0.045\\
Random           	& 0.000     & 0.024	    & 0.023	    & 0.023\\
\hline
\end{tabular}
\caption{\label{tbl:car:eucl} Retrieval of clients with a car insurance via
dot-product similarity on mSDA embeddings. We are evaluating the MAP for the 50, 2E4, 5E4 and 9E4 nearest database points to the query.}
\end{table}

\begin{figure}[!h]
    \centering
    \includegraphics[width=.5\textwidth]{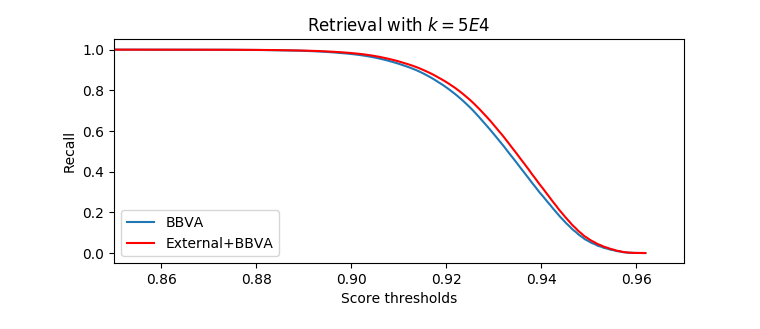}
    \caption{The recall curves for the retrieval of clients where the relevance of a client is defined by having a BBVA insurance (blue) or, additionally, with a specific external provider (red). We use the 5E4 nearest neighbors to each query client.}
    \label{fig:recall_external}
\end{figure}

We evaluate the MAP@$k$ for various values of $k$. The scores in Table \ref{tbl:car:eucl}

show that mSDA embeddings target customers more efficiently than random search, although the scores are nominally not far apart. We initially attributed this to the small fraction of relevant customers or to a weak correlation between insurance-related expenses and the other transactions. 
However, by looking at the 
blue recall curve of the retrieval problem with $k=5E4$ depicted in Figure \ref{fig:recall_external}, we realize that we are 
indeed retrieving a lot of relevant clients, thus indicating that we must be introducing 
many false positives.

Commercially, this is not an unwelcome outcome: we may be
retrieving the real targets of an effective campaign, that is those customers 
that {\em do} have a car insurance, only with a different company. This information is not immediately evident from our data which doesn't usually detail the insurance company or the kind of insurance (house, car, health etc.). Through some mining, however, we detected 
an exception to this rule by a descriptor specifying that some transactions referred to 
a car policy sold by a specific Spanish insurer, which allowed us to verify the positivity of some of those supposed false positives. This is equivalent to identifying a new set $M$ to 
define the positive labels (hence the relevance) of the database clients, consequently improving the MAPs as per Table \ref{tbl:car:eucl:mutua}.

\begin{table}[!h]
\centering
\begin{tabular}{llccccc}
\hline
Method              & MAP@50    & @2E4      & @5E4      & @9E4  \\
\hline
mSDA + dot 	        & 0.289	    & 0.108	    & 0.072	    & 0.054\\
Random           	& 0.068	    & 0.035	    & 0.035	    & 0.035\\
\hline
\end{tabular}
\caption{\label{tbl:car:eucl:mutua} Retrieval of clients with a car insurance with BBVA or
an external provider via mSDA and dot-product scoring.}
\end{table}

The red curve in Figure \ref{fig:recall_external} is the recall for this new relevance criterion. Notice that the difference with the BBVA-only insurance recall is minimal. That is, the increase in MAP can be traced back to an increase in precision, but the recall graph suggests there may still be ``false'' positives we are
mislabeling in our ground truth.

In order to spot them, we hypothesize the algorithm is retrieving all clients with car-related expenses, e.g. \textit{Highway toll}, \textit{Automobile repair shops}, \textit{Petrol}, whom we thus label manually as positives. Customers with transactions in 
these categories are highly likely to own a car, which means they likely have a 
car insurance with a different company. The MAPs for this case are 
reported in Table \ref{tbl:car:eucl:all}, showing a substantial improvement on 
all values of $k$.

\begin{table}[!h]
\centering
\begin{tabular}{llccccc}
\hline
Method              & MAP@50    & @2E4      & @5E4      & @9E4  \\
\hline
mSDA + dot 	        & 0.897	    & 0.776	    & 0.670	    & 0.536\\
Random           	& 0.068	    & 0.035	    & 0.035	    & 0.035\\
\hline
\end{tabular}
\caption{\label{tbl:car:eucl:all} Retrieval of clients with car-related expenses.}
\end{table}

\begin{figure}[!t]
    \centering
    \includegraphics[width=.35\textwidth]{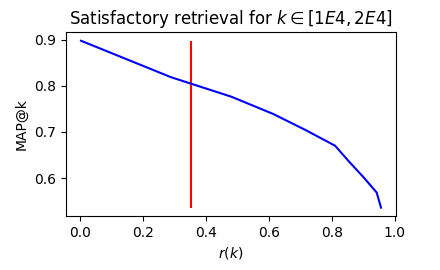}
    \caption{MAP values for the fraction of retrieved clients for each $k$. The red line
    is the proportion of positive database clients. A high MAP at the corresponding $r(k)$
    suggests most relevant clients are being retrieved.}
    \label{fig:maps_for_k}
\end{figure}

Finally, we conclude the section by arguing that restricting the retrieval to some of 
the values of $k$ which we have considered allows us to hit a sweetspot between diversity 
and precision. Call $R(k)$ the number of distinct database clients retrieved for a certain $k$ and 

\begin{equation}
    r(k) = \dfrac{R(k)}{|\mathcal{D}|}
\end{equation}

the fraction of distinct database clients retrieved when using $k$ neighbors. In Figure 
\ref{fig:maps_for_k} we have plotted the values of the MAP@$k$ against the corresponding
$r(k)$ and we have also highlighted the proportion of positive database clients. A high MAP at that $r(k)$, which is achieved for $k\in[1E4, 2E4]$, indicates
that both recall and precision are sufficiently high to be 
retrieving a large number of all the relevant clients.

This sweetspot of diversity and precison may not always be achievable, in which case it is advisable to step back and consider solutions like curating the query set or investigating whether the embedding is failing to capture the 
information we'd like it to leverage for our application (one may then resort to additional
techniques like metric learning) or, like in the case described in this section, it is capturing 
way 
more than we initially expected. This latter point, in particular, demonstrates how client2vec can offer a quick and powerful starting point for data mining and advanced analytics, as well as off-the-shelf production-ready solutions.

\section{Conclusions}

We started the work which led to this study as an attempt to develop 
an internal tool that could catalyze the data-driven decision making
for BBVA. We described how we worked towards a solution that was simple to use,
fast to deploy and integrate in our colleagues' processes and that required minimal preprocessing. Along the way, we learned that composing transactional embeddings extracted with word2vec into customer embeddings doesn't always offer an acceptable performance, while mSDAs help us capture a good deal of behavioral information. Furthermore, we highlighted how 
this information can be extracted 
even from simple, coarse transactional data.
We plan to keep expanding the client2vec library by adding new 
representations as new use cases arise, as well as by 
proactively exploring algorithms that fit its philosophy of simplicity, such as the nonlinear extension of mSDA proposed in \cite{chen2014marginalized} or metric learning to further boost the performance mSDA embeddings in client targeting.
We also intend to start leveraging temporal information about the 
transactions of a client.

\section{Acknowledgements}
We would like to thank the whole team at BBVA D\&A for guiding, inspiring and counseling us. Our thanks also go to the Smart Decision Making and Personal Finance Management teams at BBVA, whose products, data and business cases helped make this research possible.

\bibliographystyle{abbrv}
\bibliography{c2vkdd}  

\end{document}